\documentclass[smallcondensed]{svjour3}     
\smartqed  
\usepackage[utf8]{inputenc}
\usepackage[english]{babel}
\usepackage{graphicx}
\usepackage{booktabs}
\usepackage{color}
\usepackage{amssymb}

\begin{document}

\title{Analyzing Business Process Anomalies Using Autoencoders}


\subtitle{}


\author{
  Timo Nolle \and
  Stefan Luettgen \and
  Alexander Seeliger \and
  Max M\"uhlh\"auser
}


\institute{
  Timo Nolle, Stefan Luettgen, Alexander Seeliger, Max M\"uhlh\"auser \at Technische Universit\"at Darmstadt, Telecooperation Lab, Germany\\
  \email{\{nolle, seeliger, max\}@tk.tu-darmstadt.de}\\
  \email{stefan.luettgen@stud.tu-darmstadt.de}
}

\date{Received: date / Accepted: date}

\maketitle

\begin{abstract}
Businesses are naturally interested in detecting anomalies in their internal processes, because these can be indicators for fraud and inefficiencies.
Within the domain of business intelligence, classic anomaly detection is not very frequently researched.
In this paper, we propose a method, using autoencoders, for detecting and analyzing anomalies occurring in the execution of a business process.
Our method does not rely on any prior knowledge about the process and can be trained on a noisy dataset already containing the anomalies.
We demonstrate its effectiveness by evaluating it on 700 different datasets and testing its performance against three state-of-the-art anomaly detection methods.
This paper is an extension of our previous work from 2016 \cite{nolle2016unsupervised}.
Compared to the original publication we have further refined the approach in terms of performance and conducted an elaborate evaluation on more sophisticated datasets including real-life event logs from the Business Process Intelligence Challenges of 2012 and 2017.
In our experiments our approach reached an $F_1$ score of 0.87, whereas the best unaltered state-of-the-art approach reached an $F_1$ score of 0.72.
Furthermore, our approach can be used to analyze the detected anomalies in terms of which event within one execution of the process causes the anomaly.

\end{abstract}

\section{Introduction}
\label{sec:introduction}

Anomaly detection is becoming an integral part of business intelligence.
Businesses are naturally interested in detecting anomalies in their processes, because these can be indicators for inefficiencies in their process, badly trained employees, or even fraudulent behavior.
Consequently, being able to detect such anomalies is of great value, for they can have enormous impact on the economic well-being of the businesses.

More and more companies rely on process-aware information systems (PAISs) \cite{dumas2005process} to improve their processes.
This increasing number of PAISs has generated a lot of interest in the data these systems are gathering.
The log files these systems are storing can be used to extract the events executed in the process, and thereby create so called event log files.
Event logs are comprised of activities (and other miscellaneous information) that occurred during the execution of the process.
These event logs enable process analysts to explore the underlying process.
In other words, the event log consists of footprints of the process.
Consequently, it is possible to recreate the process model by evaluating its event log.
This is known as \textit{process model discovery} and is one of the main ideas in the domain of process mining \cite{van2011process}.

Process mining provides methodologies to detect anomalies in the execution of a process; e.g., by discovering the as-is process model from the event log \cite{van2004workflow} using discovery algorithms and then comparing the discovered model to a reference model.
This is known as \textit{conformance checking} \cite{rozinat2008conformance}.
Another way of detecting anomalies is to compare the event log to the reference model.
However, this approach requires the existence of such a reference model.

If no reference model is available, process mining relies on discovering a reference model from the event log itself \cite{Bezerra2009Anomaly,bezerra2013algorithms}.
These methods make use of a threshold to deal with infrequent behavior in the log, so that the discovered model is a good representation of the normal behavior of the process.
Hence, this model can be used as a reference model for the conformance check.

A key assumption in anomaly detection is that the anomalous executions occur less frequent than normal executions.
This skewed distribution can be taken advantage of when applying anomaly detection techniques.

In this paper, we propose a method for detecting anomalies in business process data.
Our method works under the following assumptions.
\begin{itemize}
    \item No prior knowledge about the process
    \item Training data already contains anomalies
    \item No reference model needed
    \item No labels needed (i.e., no knowledge about anomalies)
    \item The algorithm must detect the exact activity at which the anomaly occurred
\end{itemize}

The system must deduce the difference between normal and anomalous executions purely based on the patterns in the raw data.
Our approach is based on a special type of neural network, called an autoencoder, that is trained in an unsupervised fashion.

The main contribution of this work is the application of an autoencoder to analyze the detected anomalies in terms of which event within a sequence is anomalous as opposed to the whole sequence at once.
This can be refined further by analyzing which characteristic of the event (e.g., the executing user) is anomalous, not just the event itself.
We demonstrate that, using this approach, we can accurately identify activities that have been executed in the wrong order, skipped, or unnecessarily reworked.
Furthermore, we can detect when unauthorized users have illegally executed an activity.

To demonstrate the feasibility of our approach we compare its performance to seven state-of-the-art methods for anomaly detection.
In addition to these six methods, we also present an adaptation of one of the methods.
All methods were applied to a comprehensive set of 600 different artificial event logs featuring authentic business process anomalies as well as 100 real-life event logs coming from the Business Process Intelligence Challenge (BPIC).

In summary, the contributions of this paper are as follows.
\begin{enumerate}
    \item Novel application of autoencoders to automatically analyze anomalies in the domain of business process intelligence.
    \item Adaptation of the t-STIDE anomaly detection method from \cite{warrender1999detecting} to work with event attributes.
    \item Comprehensive evaluation of state-of-the-art anomaly detection methods in the domain of business process intelligence.
    \item Provision of a representative, labelled, set of artificial process event logs containing authentic anomalies.
\end{enumerate}

\section{Related Work}
\label{sec:related_work}

In the field of process mining \cite{van2011process} anomaly detection is not very frequently researched.
Most proposed methods work by using discovery algorithms to mine a reference model from the event log \cite{Bezerra2009Anomaly} and then using it  for conformance checking to detect anomalous behavior.
The bigger part of these methods relies on a clean dataset to work correctly.
Unfortunately, this violates our assumptions, as the data coming from the PAISs will naturally contain anomalies.

Recently there has been some research on approaches that can deal with noisy event logs.
Through the use of special discovery algorithms, that can deal with noise and infrequent behavior in the process, the approach from \cite{Bezerra2009Anomaly} can be refined to work with noisy logs \cite{bezerra2013algorithms}.
The authors in \cite{bezerra2013algorithms} give three different algorithms in their paper.
Within this work we will compare our approach to two of the proposed approaches.

A more recent publication proposes the use of likelihood graphs to analyze business process behavior \cite{bohmer2016multi}.
Specifically, the authors describe a method to extend the likelihood graph to include event attributes.
This method works both on noisy event logs and includes important characteristics of the process itself by including the event attributes.
We will also compare our method to the method from \cite{bohmer2016multi} in the evaluation section.

A review of classic anomaly detection methodology can be found in \cite{pimentel2014review}.
Here, the authors describe and compare many methods that have been proposed over the last decades.
Another elaborate summary on anomaly detection in discrete sequences is given by Chandola in \cite{chandola2012survey}.
The authors differentiate between five different basic methods for novelty detection: probabilistic, distance-based, reconstruction-based, domain-based, and information-theoretic novelty detection.

Probabilistic approaches try to estimate the probability distribution of the normal class, and thus can detect anomalies as they were sampled from a different distribution.
In speech recognition \cite{juang1991hidden}, hidden Markov models (HMMs) \cite{rabiner1986introduction,rabiner1989tutorial} are a popular choice for modeling sequential data.
HMMs can also be used for anomaly detection as shown in \cite{warrender1999detecting} and \cite{Jain2012HMMintrusion}, where they are used successfully for system intrusion detection.
However, as Chandola pointed out in \cite{chandola2008report}, the performance of such HMMs strongly depends on the fact that the raw data can be sufficiently modeled by a Markov process.

Another important probabilistic technique is the sliding window approach as proposed in \cite{forrest1996sense}, where it is used for intrusion detection.
In window based anomaly detection, every window of a sequence is assigned an anomaly score.
Then the anomaly score of the sequence can be inferred by aggregating the window anomaly scores.
Recently, Wressnegger et\,al. used this approach for intrusion detection and give an elaborate evaluation in \cite{wressnegger2013acloselook}.
While being inexpensive and easy to implement, sliding window approaches show a robust performance in finding anomalies in sequential data, especially within short regions of the data \cite{chandola2012survey}.

Distance-based novelty detection does not require a cleaned dataset, yet it is only partly applicable for process traces, as anomalous traces are usually very similar to normal ones.
A popular distance-based approach is the one-class support vector machine (OC-SVM).
Sch\"olkopf et\,al. \cite{scholkopf1999support} first used support vector machines \cite{cortes1995support} for anomaly detection.
Tax, in his PhD thesis \cite{Tax2001oneClass}, gives a sophisticated overview over one-class classification methods, also mentioning the OC-SVM.
OC-SVMs have shown to be successful in the field of intrusion detection as demonstrated by \cite{Heller03oneclass}.

Reconstruction-based novelty detection (e.g., neural networks) is similar to the aforementioned approaches in \cite{hawkins2002outlier,Japkowicz2001}.
However, training a neural network usually also requires a cleaned dataset.
Nevertheless, we will show that our approach works on the noisy dataset by taking advantage of the skewed distribution of normal data and anomalies, as demonstrated in \cite{eskin2000anomaly}.

Domain-based novelty detection requires domain knowledge, which violates our assumption of no prior knowledge about the process.
Information-theoretic novelty detection defines anomalies as the examples that most influence an information measure (e.g., entropy) on the whole dataset.
Iteratively removing the data with the highest impact will yield a cleaned dataset, and thus a set of anomalies.

The approach within this paper is highly influenced by the works in \cite{dong2016threaded,hawkins2002outlier,Japkowicz2001}, in which they propose the use of replicator neural networks \cite{hecht1995replicator} for anomaly detection, i.e., networks that reproduce their input, which are based on the idea of autoencoders from \cite{hinton1989connectionist}.
Autoassociative neural network encoders use a similar concept and have been used to model the nominal behavior of complex systems \cite{thompson2002implicit}.
They have also been used for residual generation in \cite{diaz2002residual}, demonstrating that these models can also model behavior not directly observed in the training data, which increases generalization.
A comprehensive study of replicator neural networks for outlier detection can be found in \cite{williams2002comparative}.
The approaches from \cite{diaz2002residual,dong2016threaded,hawkins2002outlier,Japkowicz2001,thompson2002implicit}, however, do not work well with variable length input.
In our approach, we address this problem by using a padding technique.
We opted to use a neural network based approach, for recent achievements in machine translation and natural language processing indicate that neural networks are an excellent choice when modeling sequential data \cite{dai2015semi,lecun2015deep}.

The main distinction between all other methods and the proposed approach is that it can be used to identify which exact event and furthermore which attribute characteristic is the cause of the anomaly.
The only other approach that can deal with event attributes is the method from \cite{bohmer2016multi}.
However, it can not deal with long-term dependencies, because it works on a general likelihood graph, which disregards the past events when calculating the probability of an event occurring at a specific point in the process.
Our approach can deal both with the attributes and with non-local dependencies in the logs.

\section{Dataset}
\label{sec:dataset}

PAISs keep a record of almost everything that happened during the execution of a business process.
This information can be extracted from the systems in form of an event log.
Event logs are the most common data structure when working with process data from PAISs, especially in the field of process mining.

\subsection{Event logs}
An event log consists of traces, each consisting of the activities that have been executed.
Table~\ref{tab:event_log} shows an excerpt of such an event log.
In this case, it is representative for the execution of a procurement process.
Notice that an event log must consist of at least three columns: a trace ID, to uniquely assign an executed activity to a trace; a timestamp, to order the activities within a trace; and an activity label, to distinguish the different activities.
Optionally, the event log can contain so called event attributes.
In the example event log from Tab.~\ref{tab:event_log}, the user column is such an event attribute, indicating which user has executed the respective activity.

\begin{table}
  \centering
  \caption{Example event log of a procurement process}
  \label{tab:event_log}
  \begin{tabular}{clll}
    \toprule
    Trace ID & Timestamp    & Activity      & User      \\
    \midrule
    1 & 2015-03-21 12:38:39 & PR Created    & Roy       \\
    1 & 2015-03-28 07:09:26 & PR Released   & Earl      \\
    1 & 2015-04-07 22:36:15 & PO Created    & James     \\
    1 & 2015-04-08 22:12:08 & PO Released   & Roy       \\
    1 & 2015-04-21 16:59:49 & Goods Receipt & Ryan      \\
    \midrule
    2 & 2015-05-14 11:31:53 & SC Created    & Marilyn   \\
    2 & 2015-05-21 09:21:26 & SC Purchased  & Emily     \\
    2 & 2015-05-28 18:48:27 & SC Approved   & Roy       \\
    2 & 2015-06-01 04:43:08 & PO Created    & Johnny    \\
    \bottomrule
  \end{tabular}
\end{table}

\subsection{Process model generation}
To create a test setting for our approach we randomly generated process models and then sampled event logs from them.
The process models were generated using PLG2 \cite{Burattin2015Plg2}, a process simulation and randomization tool.
Each process model has a different complexity, with regard to the number of possible activities and the branching factors (i.e., out-degrees).
The complexity of a process model can also be measured by the number of possible variants.
A variant is a valid path through the complete process model from a valid start activity to a valid end activity.
Table~\ref{tbl:process_models} shows the process models with their corresponding complexities.
Note that the \textit{Wide} process model was specifically generated to evaluate the approach on a dataset that has low complexity in terms of the number of variants, but a high branching factor.

\begin{table}[t]
  \centering
  \caption{Overview over the 5 different randomly generated process models and the P2P process}
  \begin{tabular}{lrrrrr}
    \toprule
    Model  & \#Nodes & \#Edges & \#Variants & Max length & $\varnothing$ Out-degree \\
    \midrule
    P2P    & 14      & 16      &  6         &  9         & 1.14                     \\
    Small  & 22      & 26      &  6         & 10         & 1.18                     \\
    Medium & 34      & 48      & 25         &  8         & 1.41                     \\
    Large  & 44      & 56      & 28         & 12         & 1.27                     \\
    Huge   & 56      & 75      & 39         & 11         & 1.34                     \\
    Wide   & 36      & 53      & 19         &  7         & 1.47                     \\
    \bottomrule
  \end{tabular}
  \label{tbl:process_models}
\end{table}

Now, we generated authentic event logs from these process models by randomly sampling variants of the process with replacement.
In real process models these variants are not equally distributed.
Therefore, we randomly generated a distribution for the variants each time we were sampling an event log.
These probabilities were sampled from a normal distribution with $\mu = 1$ and $\sigma = 0.2$, and then normalized so they sum up to 1.
Furthermore, we randomly generated a set of users in the process (between 10 and 30 different users per process).
Then we sampled subsets of the user set for each activity, denoting which users are permitted to execute the activity.
The number of possible users per activity lies between 1 and 5.
After computing all variants, we also introduced a long-term dependency for the user variable in each variant at random.
Therefore, we randomly chose two activities in each variant that must be executed by the same user.

\subsection{Example process}
\begin{figure}[t]
    \centering
    \includegraphics[width=0.6\linewidth]{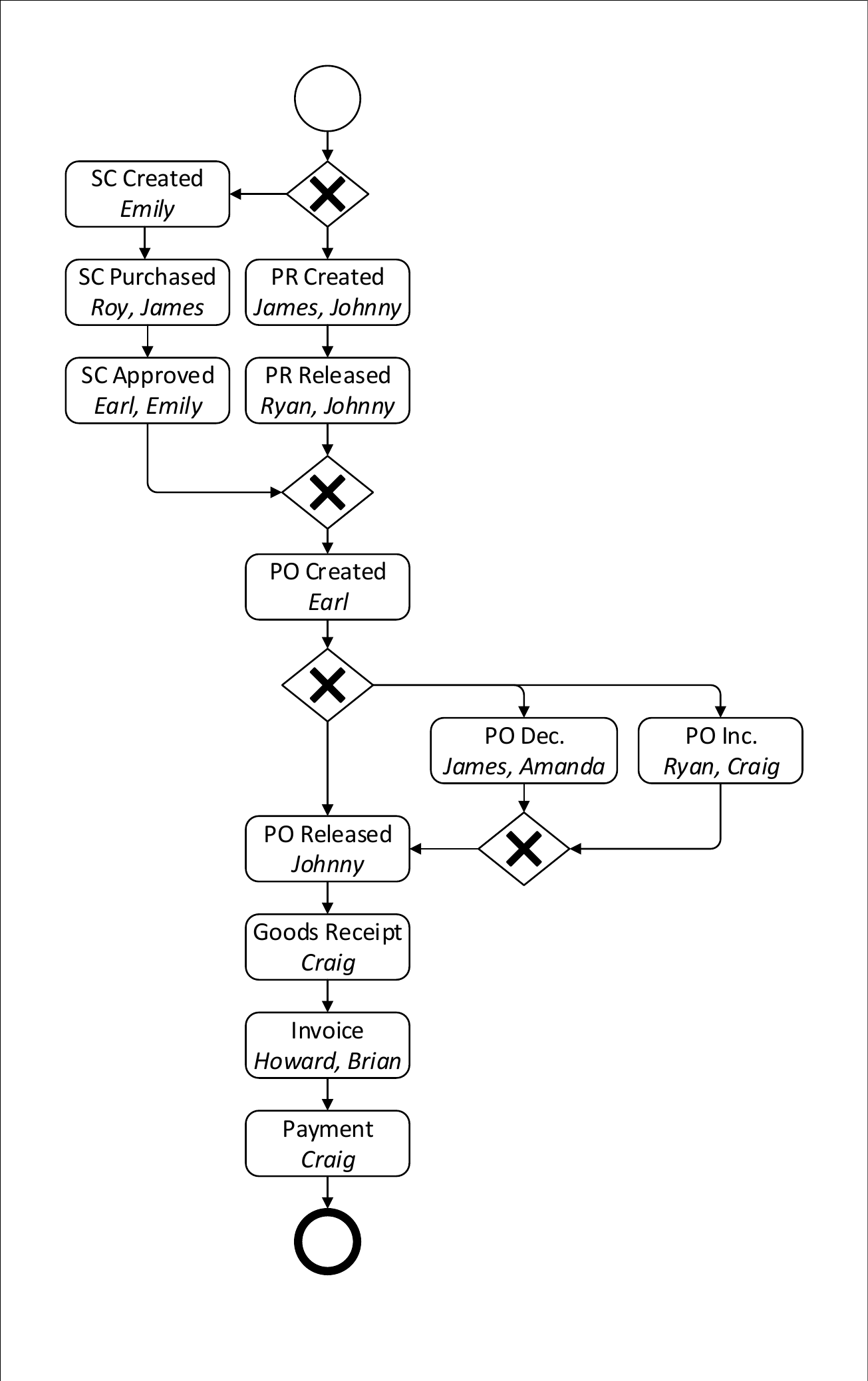}
    \caption{BPMN model of a simplified purchase to pay process; the italic names represent the users allowed to execute that activity}
    \label{fig:process_model}
\end{figure}

In addition to the five randomly generated models, we also used a simplified version of a purchase to pay (P2P) process model as is depicted by the BPMN model in Fig.~\ref{fig:process_model}.
This model was mainly created for purposes of evaluation, as it features interpretable activity names unlike the randomly generated models.
The resulting event log for the P2P model was generated in the same fashion as those of the randomly generated models using the same parameters as mentioned above.
Notice the possible users for each activity as indicated by the italic names in Fig.~\ref{fig:process_model}.

\subsection{Anomalies}
To introduce noise into the event logs we randomly applied mutations to a fixed percentage of the traces in the event log.
These mutations represent the anomalies in the data.
Each trace can be affected by one of the following five anomalies (we will use their respective names from now on):

\begin{enumerate}
    \item \textit{Skipping}: A necessary activity has not been executed,
    \item \textit{Switching}: Two consecutive events have been executed in the wrong order,
    \item \textit{Reworking}: An activity has been executed twice in a row,
    \item \textit{Incorrect user}: A user has executed an activity to which he was not permitted,
    \item \textit{Incorrect LTD}: The wrong user has executed the long-term dependent activity.
\end{enumerate}

Compared to our work in \cite{nolle2016unsupervised}, we have added two more anomalies that we found occur very frequently in real-life scenarios.
A classic problem in real-life business processes is the segregation of duty.
For example, a user that approves a purchase order must not be the same user that has initially created it.
Many anomalies in real-life processes are related to the users executing the events, which is why we included this event attribute here.

Our way of generating the artificial event logs is very similar to the methods of Bezerra \cite{bezerra2013algorithms} and B\"ohmer \cite{bohmer2016multi}.
One difference is, that we also introduce anomalies affecting event attributes.
We will make these datasets, the generation algorithm, and our implementation of the algorithm publicly available.
For more information on this, please consider contacting the corresponding author.

For each process model, we randomly generated a set of permitted users for each activity.
We did this ten times, resulting in 60 different process models.
For each of these 60 process models, we then generated 10 event logs, each featuring a different percentage of anomalies and a random variant distribution.
The percentage of anomalous traces in the training log ranged from 10\%, 20\%, up to 100\%.
That is, we generated training logs containing 10\% anomalies and 90\% normal traces, as well as logs with 80\% anomalies and 20\% normal traces, and so on up to a log which entirely consists of anomalies, i.e., 100\%.
In total, we generated 600 different artificial event logs.
Each event log consisted of 12\,500 traces.
For each event log we created a separate test event log containing 2\,500 traces featuring the same variant distribution and users.

\subsection{Real-life event logs}
In addition to the artificial event logs we also generated training and test event logs from the public datasets of the Business Process Intelligence Challenge 2012\footnote{http://www.win.tue.nl/bpi/doku.php?id=2012:challenge} and 2017\footnote{http://www.win.tue.nl/bpi/doku.php?id=2017:challenge}, which we will refer to as BPIC12 and BPIC17 respectively.
BPIC17 is an updated version of BPIC12, representing the same loan application process.
However, BPIC17 contains data from the last 5 years, after the company has introduced a new workflow system.

Similarly to the artificial logs, we used the event logs as a basis and randomly applied anomalies to a fixed percentage of traces in the logs.
As these logs did not feature a user attribute we did not include the \textit{Incorrect user} and \textit{Incorrect LTD} anomalies.
For BPIC12 and BPIC17 we generated training sets featuring between 10\% and 100\% anomalies, as was done for the artificial logs.
We also generated separate test sets for both logs, resulting in 100 real-life training event logs with artificial anomalies.

\section{Method}
\label{sec:method}

Recently, artificial neural networks have gotten a lot of attention by outclassing the state-of-the-art methods in many domains such as object recognition in images \cite{krizhevsky2012imagenet} or machine translation \cite{bahdanau2014neural}.
Before we introduce our method, we first want to give a brief overview over the neural network architecture we employed.

A feed-forward neural network consists of multiple layers, each containing many neurons.
Every neuron in one layer is connected to all neurons in the preceding and succeeding layers.
These connections have weights attached to them, which can be used to control the impact a neuron in one layer has on the activation of a neuron in the next layer.
To calculate the output of a neuron we apply a non-linear activation function (a popular choice is the rectifier function $f(x) = \max(0, x)$ \cite{nair2010rectified}) to the sum over all outputs of the neurons in the previous layer times their respective connection weights.
The initialization of these weights is important, as pointed out in \cite{glorot2010understanding}, for no two weights within one layer must be initialized to the same value.
Then, the back-propagation algorithm \cite{rumelhart1988learning} is used to iteratively tune the weights, so that the neural network produces the desired output, or a close enough approximation of it.

In a classification setting, the desired output of the neural network is the class label.
However, a neural network can also be trained without the use of class labels.
One such type of neural network is called an autoencoder.
Instead of using class labels, we are using the original input as the target output when training the autoencoder.
Obviously, a neural network, if given enough capacity and time, can simply learn the identity function of all examples in the training set.
To overcome this issue, some kind of capacity limitation is needed.
This can be done by forcing one of the autoencoder's hidden layers to be narrow (i.e., narrower than the input dimension), thereby not allowing the autoencoder to learn the identity function.
Another common way of limiting the capacity is to distribute additive Gaussian noise over the input vector of the autoencoder.
Thus, the autoencoder---even if repeatedly trained on the same trace---will always receive a different input.
We use a combination of both these strategies for our method.

\subsection{Setup}
To train an autoencoder on the generated event logs, we first must transform them.
The first step is to encode each activity and user using a one-hot encoding.
Each activity is encoded by an $n$-dimensional vector, where $n$ is the number of different activities encountered in the event log.
To encode one activity, we simply set the corresponding dimension of the one-hot vector to a fixed value of one, while setting all the other dimensions to zero.
We use the same method to encode the user event attribute.
This results in a one-hot vector for the activity and another for the user for each event in a trace.
Now we combine these vectors by concatenating them into one vector.
If the activity vectors are $a_1, a_2, ..., a_n$ and the respective user vectors are $u_1, u_2, ..., u_n$, the resulting vector will be $a_1 \Vert u_1 \Vert a_2 \Vert u_2 \Vert ... \Vert a_n \Vert u_n$, where $\Vert$ denotes concatenation.

Note that another option of dealing with variable size traces is dividing the traces into subsequences of equal size (n-grams).
However, using n-grams of events loses the connection between distant events, if the n-gram size is too narrow.
Consequently, the system is unable accurately model long-term dependencies between events.
Therefore, we chose to use the one-hot encoding method.

Because feed-forward neural networks have a fixed size input, we must apply one more step of pre-processing.
To force all encoded trace vectors to have the same size we pad all vectors with zeros, so each vector has the same size as the longest vector (i.e., the longest trace) in the event log.

Suppose an event log consists of 10 different activities, 20 different users, and the maximum length of all traces in the event log is 12.
The longest trace within the event log will have a size of $(10 + 20) \cdot 12 = 360$.
Therefore, we must pad all shorter vectors with zeros so they reach size 360.

\begin{figure}[t]
    \centering
    \includegraphics[width=0.8\linewidth]{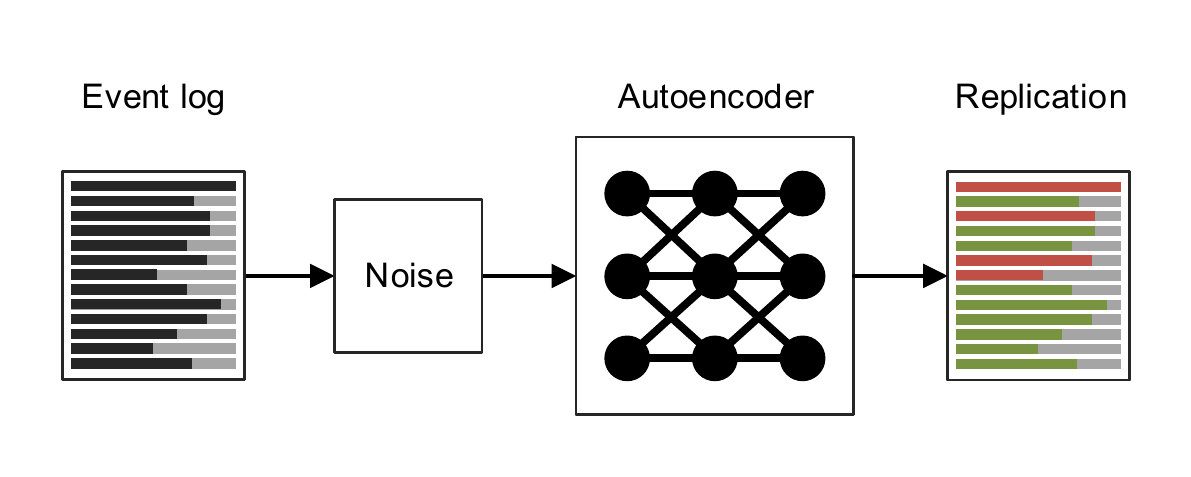}
    \caption{Autoencoder is trained to replicate the traces in the event log after the addition of Gaussian noise}
    \label{fig:training}
\end{figure}

Using the one-hot encoded event log we can train the autoencoder with the back-propagation algorithm \cite{rumelhart1988learning}, using the event log both as the input and the label.
Figure~\ref{fig:training} shows a simplified version of the architecture.
The special noise layer adds Gaussian noise before feeding the input into the autoencoder.
This layer is only active during training.
Now the autoencoder is trained to reproduce its input, that is, to minimize the mean squared error between the input and its output.

We trained on mini batches of size 50 for 200 epochs, allowing early stopping when the loss on the validation set did not decrease within the last 10 epochs.
We used the Adam optimizer \cite{kingma2014adam}, which utilizes the momentum technique \cite{sutskever2013importance}.
We set the optimizer parameters to $\beta_1=0.9$, $\beta_2=0.99$ and $\epsilon=10^{-8}$.
The learning rate was set to $0.001$ initially, and was scaled by a factor of $0.1$ when the validation loss did not improve within the last $5$ epochs.
Additionally, we used a dropout of $0.5$ between all layers, as suggested in \cite{srivastava2014dropout}; the additive noise applied to the input was sampled from a Gaussian distribution with $\mu=0$ and $\sigma=0.1$.
Each autoencoder consists of an input and an output layer with linear units, and $2$ hidden layers with rectified linear units.
These training parameters were used for each of the different event logs, but the size of the hidden layer was adapted depending on the event log, i.e., the number of neurons in the hidden layer was set to be half the size of the input layer.
For the real-life BPIC event logs we only used 1 hidden layer.

\subsection{Classifying traces}
After training the autoencoder, it can be used to reproduce the traces in the test event logs, but without applying the noise.
Now, we can measure the mean squared error between the input vector and the output vector to detect anomalies in the event log.
Because the distribution of normal traces and anomalous traces in the event log is one sided, we can assume that the autoencoder will reproduce the normal traces with less reproduction error than the anomalies.
Therefore, we can define a threshold $\tau$, where if the reproduction error of a trace succeeds this threshold $\tau$, we consider it an anomaly.
To set the threshold we use the mean reproduction error over the training dataset and apply a scaling factor $\alpha$.
We define the threshold as in Equation~\ref{eq:threshold}, where $e_i$ is the reproduction error for trace $i$, and $n$ the number of traces in the dataset.
\begin{equation}
\label{eq:threshold}
    \tau = \frac{\alpha}{n} \sum_{i=1}^n e_i
\end{equation}

\subsection{Classifying events and attributes}
We have described how to detect anomalous traces in the event log; now we want to refine this method.
Not only can we detect that a trace is anomalous, but also which event in the trace influences the reproduction error the most.
Hence, we must change our calculation of the reproduction error from trace based to event based.
Up until now, we calculated the reproduction error as the mean squared error between the entire one-hot encoded input and output sequence of the autoencoder.
However, we can also consider the mean squared error for every event in the sequence separately.
Furthermore, we can also compute the error for each activity and user separately.

Let us consider the example input vector $i$ from Equation~\ref{eq:input_vector}.
We can divide the vector into the corresponding subvectors, as indicated by the curly braces.
This gives us $a_1, u_1, a_2, u_2, ..., a_n, u_n$.
Now we can split the reproduced version of $i$ (i.e., the output vector) identically, obtaining $\hat{a}_1, \hat{u}_1, \hat{a}_2, \hat{u}_2, ..., \hat{a}_3, \hat{u}_3$.
\begin{equation}
    \label{eq:input_vector}
    i = [ ~ \underbrace{00001}_{a_1} ~ \underbrace{0100}_{u_1} ~ \underbrace{10000}_{a_2} ~ \underbrace{0010}_{u_2} ~ ... ~ \underbrace{01000}_{a_n} ~ \underbrace{0100}_{u_n} ~ ]
\end{equation}
The error $E$ for an activity vector $a_i$ is then given by the mean squared error between $a_i$ and $\hat{a}_i$.
For a user vector $u_i$ the method works analogously.
Thus, we can compute the error for all activity vectors and all user vectors over the whole dataset.
Notice that this works for any number of event attributes.

The benefit is that we can distinguish between activity related anomalies and user related anomalies.
We will elaborate on this in the evaluation section below.

\section{Evaluation}
\label{sec:evaluation}

We evaluated the autoencoder approach (DAE) on all 700 event logs and compared it to state-of-the-art anomaly detection methods mentioned \cite{chandola2012survey}.
Namely: a sliding window approach named t-STIDE \cite{warrender1999detecting}; the one-class SVM approach (OC-SVM); and the Markovian approach using a hidden Markov model (HMM) \cite{warrender1999detecting}.
In addition to that, we also compared our approach to two approaches proposed in \cite{bezerra2013algorithms}, the Naive algorithm and the Sampling algorithm.
Lastly, we compared our approach to the most recent approach proposed in \cite{bohmer2016multi}, using an extended likelihood graph (Likelihood).
As a baseline we provide the results of a random classifier.

For the OC-SVM we relied on the implementation of the scikit-learn package for Python \cite{scikit-learn} using an RBF kernel of degree 3 and a $\nu = 0.6$.
The HMM approach was implemented using the hmmlearn package for Python.
We implemented the t-STIDE algorithm ourselves using a window size $k = 4$.
The hyperparameters for both approaches were optimized using grid search.
The Naive, Sampling, and Likelihood methods were implemented as described in the original papers.

At last, we used our own implementation of the t-STIDE method which we will refer to as t-STIDE+.
The classic t-STIDE approach only takes into account the activities of an event log, but not the attributes.
To make use of the attributes we must adapt the original method.

A window of size $k$ is a tuple of $k$ events, where each event consists of a tuple of the activity name $a$ and the corresponding user $u$.
Let us consider an example window size of three.
A window $w$ is defined as $w = \{(a_1, u_1), (a_2, u_2), (a_3, u_3)\}$.
The approach works by employing a frequency analysis over all windows of size $k$ in the training set, and then comparing the relative frequencies of all windows in the test set to the corresponding ones from the training set.
Whenever a window's relative frequency is significantly lower than its frequency in the training set, the trace containing this window is considered an anomaly.
We evaluated the t-STIDE and the t-STIDE+ approach on all datasets and for all feasible choices of $k$, i.e., $k$ was chosen to lie between $2$ and the maximum trace length in the dataset.
The evaluation showed that $k=4$ performed the best for both approaches.

We used the threshold technique from Equation~\ref{eq:threshold} for all approaches except the OC-SVM, for the scikit-learn implementation automatically optimizes the threshold.
For the other approaches, we optimized the scaling factor $\alpha$ by an exhaustive grid search.
One requirement when setting $\alpha$ was that $\alpha$ must be the same for all event logs, i.e., we strive for a general setting of $\alpha$.

We also considered isolation forests \cite{liu2008isolation,liu2012isolation} for our experiments; however, this approach relies on setting a contamination parameter indicating the noise level of the data, rendering the approach unusable as we assume no prior knowledge about the noise level.

We evaluated the 9 methods on all 600 artificial, as well as the 100 real-life event logs.
In total, we evaluated $6\,300$ models.

\subsection{Experiment results}
\begin{figure}[t]
    \includegraphics[width=1.0\textwidth]{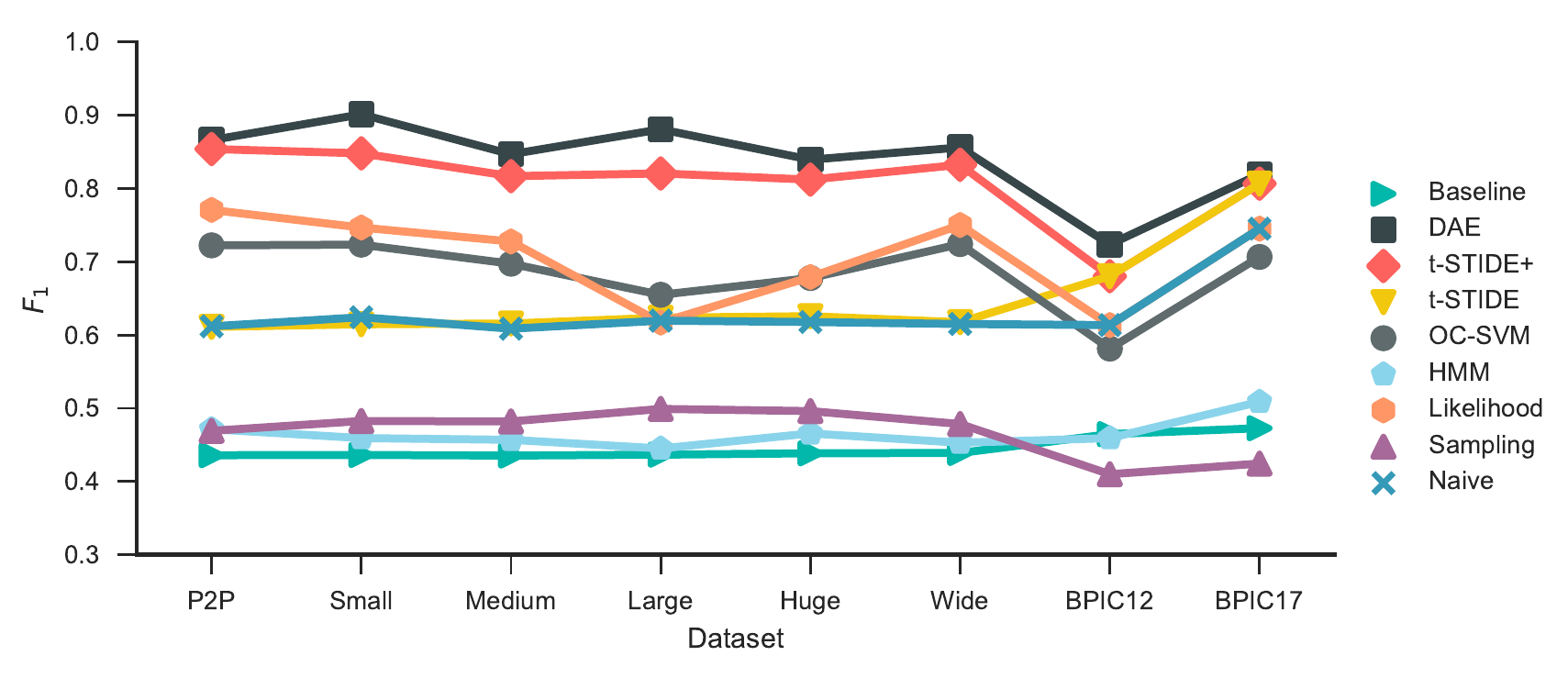}
    \caption{$F_1$ score by process model and method}
    \label{fig:f1_by_model}
\end{figure}

\begin{table}[t]
    \centering
    \caption{Results of the experiments for all evaluated methods for each process model; best results are shown in bold typeface}
    \label{tab:f1_by_model}
    \resizebox{\columnwidth}{!}{%
        \begin{tabular}{lcccccc}
            \toprule
                                                  & P2P             & Small           & Medium          & Large           & Huge            & Wide            \\
            \midrule
            Baseline                              & 0.44 $\pm$ 0.01 & 0.44 $\pm$ 0.01 & 0.44 $\pm$ 0.01 & 0.44 $\pm$ 0.01 & 0.44 $\pm$ 0.01 & 0.44 $\pm$ 0.01 \\
            HMM~\cite{warrender1999detecting}     & 0.47 $\pm$ 0.02 & 0.46 $\pm$ 0.01 & 0.46 $\pm$ 0.01 & 0.45 $\pm$ 0.02 & 0.47 $\pm$ 0.01 & 0.45 $\pm$ 0.02 \\
            OC-SVM~\cite{scholkopf1999support}    & 0.72 $\pm$ 0.06 & 0.72 $\pm$ 0.05 & 0.70 $\pm$ 0.05 & 0.65 $\pm$ 0.04 & 0.68 $\pm$ 0.05 & 0.72 $\pm$ 0.05 \\
            Naive~\cite{bezerra2013algorithms}    & 0.61 $\pm$ 0.01 & 0.62 $\pm$ 0.01 & 0.61 $\pm$ 0.03 & 0.62 $\pm$ 0.01 & 0.62 $\pm$ 0.02 & 0.62 $\pm$ 0.02 \\
            Sampling~\cite{bezerra2013algorithms} & 0.47 $\pm$ 0.03 & 0.48 $\pm$ 0.05 & 0.48 $\pm$ 0.06 & 0.50 $\pm$ 0.06 & 0.50 $\pm$ 0.06 & 0.48 $\pm$ 0.05 \\
            t-STIDE~\cite{warrender1999detecting} & 0.61 $\pm$ 0.01 & 0.61 $\pm$ 0.03 & 0.62 $\pm$ 0.02 & 0.62 $\pm$ 0.01 & 0.63 $\pm$ 0.01 & 0.62 $\pm$ 0.02 \\
            Likelihood~\cite{bohmer2016multi}     & 0.77 $\pm$ 0.17 & 0.75 $\pm$ 0.14 & 0.73 $\pm$ 0.15 & 0.62 $\pm$ 0.10 & 0.68 $\pm$ 0.11 & 0.75 $\pm$ 0.17 \\
            \midrule
            t-STIDE+                              & 0.85 $\pm$ 0.09 & 0.85 $\pm$ 0.08 & 0.82 $\pm$ 0.09 & 0.82 $\pm$ 0.07 & 0.81 $\pm$ 0.05 & 0.83 $\pm$ 0.11 \\
            DAE                                   & \textbf{0.87 $\pm$ 0.09} & \textbf{0.90 $\pm$ 0.07} & \textbf{0.85 $\pm$ 0.08} & \textbf{0.88 $\pm$ 0.07} & \textbf{0.84 $\pm$ 0.08} & \textbf{0.86 $\pm$ 0.08} \\
            \bottomrule
        \end{tabular}
    }
\end{table}

\begin{table}[t]
    \centering
    \caption{Results on the BPIC event logs; best results are shown in bold typeface}
    \label{tab:f1_by_model_real}
    \begin{tabular}{lcc}
        \toprule
                                              & BPIC12          & BPIC17          \\
        \midrule
        Baseline                              & 0.46 $\pm$ 0.01 & 0.47 $\pm$ 0.01 \\
        HMM~\cite{warrender1999detecting}     & 0.46 $\pm$ 0.00 & 0.51 $\pm$ 0.00 \\
        OC-SVM~\cite{scholkopf1999support}    & 0.58 $\pm$ 0.07 & 0.71 $\pm$ 0.04 \\
        Naive~\cite{bezerra2013algorithms}    & 0.61 $\pm$ 0.12 & 0.75 $\pm$ 0.12 \\
        Sampling~\cite{bezerra2013algorithms} & 0.41 $\pm$ 0.00 & 0.42 $\pm$ 0.00 \\
        t-STIDE~\cite{warrender1999detecting} & 0.68 $\pm$ 0.14 & 0.81 $\pm$ 0.02 \\
        Likelihood~\cite{bohmer2016multi}     & 0.61 $\pm$ 0.12 & 0.75 $\pm$ 0.12 \\
        \midrule
        t-STIDE+                              & 0.68 $\pm$ 0.14 & 0.81 $\pm$ 0.02 \\
        DAE                                   & \textbf{0.72 $\pm$ 0.08} & \textbf{0.82 $\pm$ 0.05} \\
        \bottomrule
    \end{tabular}
\end{table}

Figure~\ref{fig:f1_by_model} shows the $F_1$ score of all methods for each process model.
The $F_1$ score per model was calculated using the macro average for each model.
Then all $F_1$ scores were averaged over all models for the corresponding process model.
A more detailed evaluation is given in Tab.~\ref{tab:f1_by_model} and Table~\ref{tab:f1_by_model_real}, which show the $F_1$ scores and their standard deviation for each process model, best results being shown in bold typeset.
Notice that the DAE approach performs best in all settings, closely followed by t-STIDE+, whereas the other approaches perform significantly worse.
Another interesting point is that the HMM approach performs no better than the random baseline, which supports Chandola's claim that HMMs are not a good method for anomaly detection in sequential data \cite{chandola2008report}.
Also the Sampling approach performs only slightly better than chance.
However, this is due to the fact that we average all results over all training sets including training event logs with higher share of anomalies.
In Fig.~\ref{fig:f1_by_noise} we can see that the Sampling method works only for low noise levels.

Overall, we can conclude that the DAE performs better than the state-of-the-art methods in all of our test settings.

\subsection{The impact of the noise level}
\begin{figure}[t]
  \centering
  \includegraphics[width=0.9\textwidth]{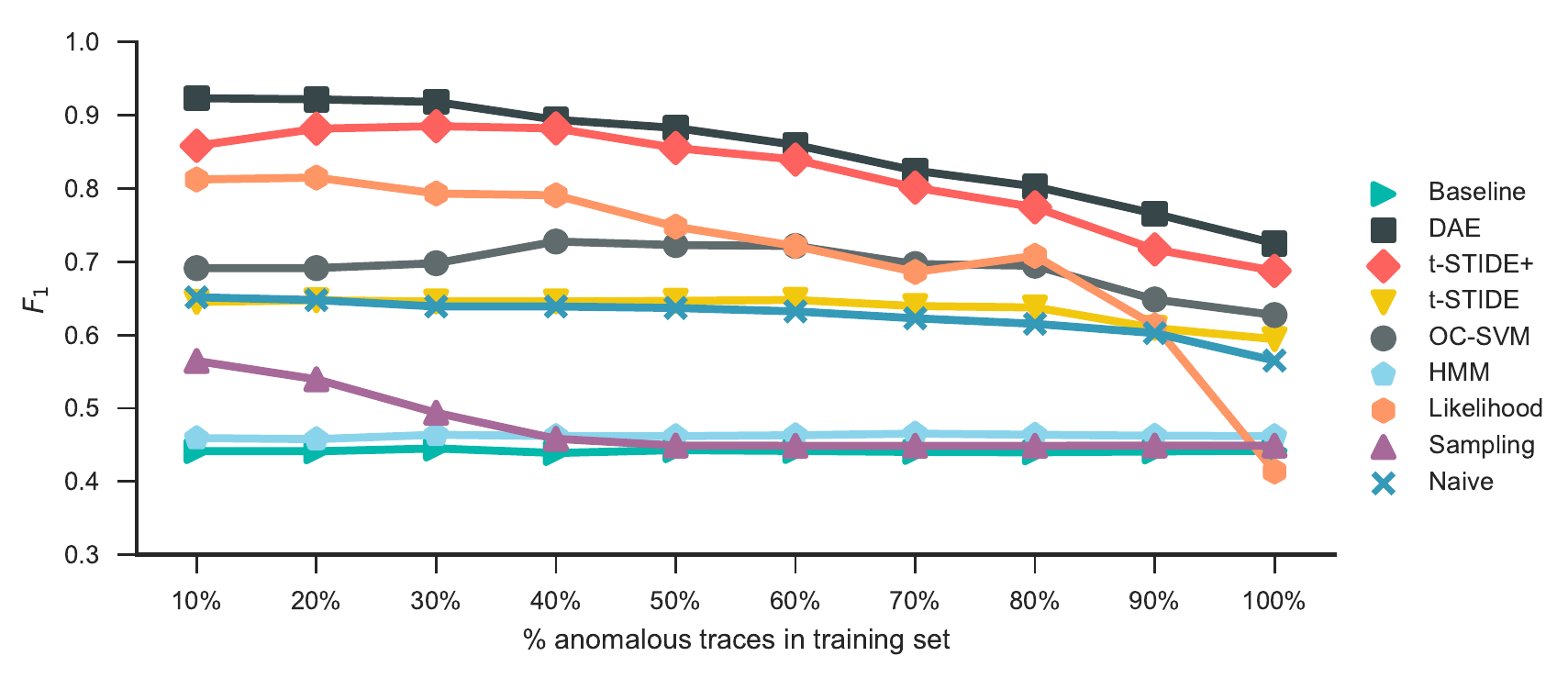}
  \caption{$F_1$ score by percentage of anomalous traces in the training set}
  \label{fig:f1_by_noise}
\end{figure}

As described before, we used different noise levels when generating the datasets, by generating training sets which included between 10\% and 100\% anomalous traces.
We use the word noise to refer to the share of traces in the training set which are anomalous.
Notice that an anomalous trace still contains normal subsequences of events.
Only a small part of the trace is affected by the anomaly in our test settings.
Hence, there is still normal behavior present in parts of each trace, even when each trace has been affected by an anomaly, as in the 100\% case.
We specifically included these harsh noise levels to test the different approaches on their ability to generalize.
We want to point out, however, that noise levels greater than 50\% are extremely unlikely in real-world settings.

One can also argue that a noise level greater than 50\% is illogical, because the classification task just gets inverted; hence, the anomaly class becomes the normal class.
This is not true for the same reason as before.
As we are dealing with sequential data and many different events in sequence (i.e., a trace) are assigned one label, there are still events that carry information about the normal behavior of the process.
And in most cases the normal events in an anomalous trace, still overpower the anomalous ones.
Hence, a noise level of 60\% is not the same as a noise level of 40\% with classes inverted.

Figure~\ref{fig:f1_by_noise} shows the $F_1$ score for all methods for the different noise levels.
Again, we find that the DAE outperforms the other approach at all noise levels, again closely followed by t-STIDE+.

Notice that the DAE still performs remarkably well, even when trained on the 100\% training set.
This is due to its ability to generalize over multiple traces.
The t-STIDE approaches can also generalize over multiple traces, because they classify based on windows; and the windows itself can contain a completely valid sequence of events.
These approaches can learn what a normal trace ought to look like, by combining the knowledge they gathered of normal subsequences over multiple traces.
For the t-STIDE approaches this is obvious, as the window size is usually smaller than the trace is long; hence, it is only trained on subsequences in the first place.
The DAE, on the other hand, is trained on the whole trace at once, which makes this level of generalization much more remarkable and unique among all the approaches.

\subsection{Interpreting the anomalies}
\begin{figure}[t]
  \centering
  \includegraphics[width=1.0\linewidth]{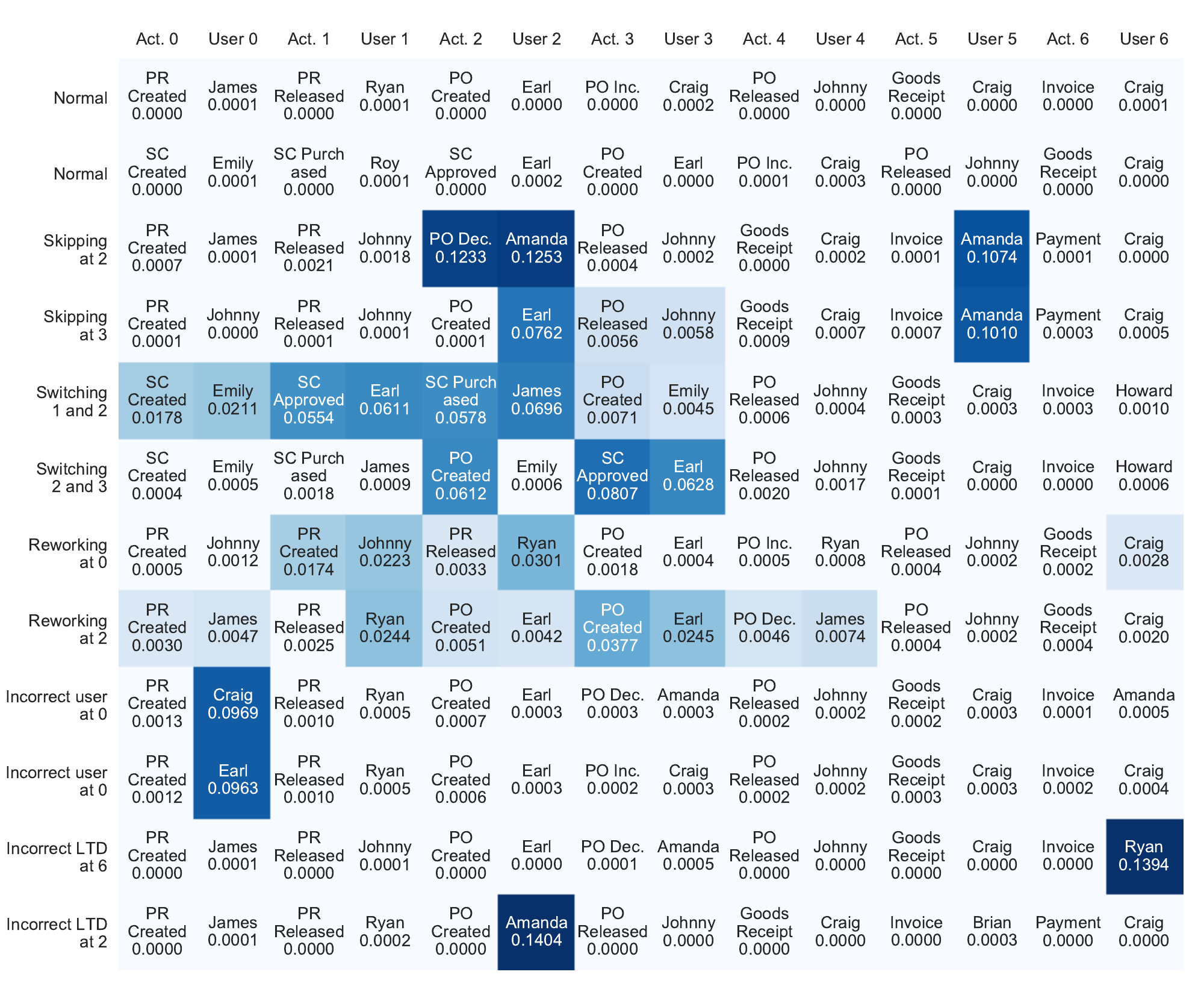}
  \caption{DAE error heatmap, trained on a P2P event log with 10\% anomalous traces}
  \label{fig:heatmap}
\end{figure}

An interesting feature of the DAE approach is that it can be used to detect not only anomalous traces, but also which event or which event attribute has influenced the reproduction error the most.
This can be done by computing the reproduction error for each event attribute separately, as described earlier.
Figure~\ref{fig:heatmap} shows 12 example traces of the P2P test dataset for a DAE trained on a training set with 10\% anomalous traces.
For clarity, we only show the first 6 events omitting the remaining events.
The cells are colored according to the reproduction error; the higher the error the darker the color.

As we can clearly see, it is never the whole trace that leads to a high reproduction error.
The DAE succeeds to reproduce the normal parts of the traces quite well, whereas it fails to reproduce the anomalous parts.
For example, the first two \textit{Normal} traces are reproduced with almost no error at all, which is exactly what we expected.
Let us now look at the four examples at the bottom (\textit{Incorrect user} and \textit{Incorrect LTD}).
The DAE is remarkably good at detecting incorrect users.
Neither Craig, nor Earl, are permitted to execute the activity \textit{PR Created} (cf.~Fig.~\ref{fig:process_model}).
Detecting \textit{Incorrect LTD} works just as fine.

Moving to the three anomalies from the original paper, we want to recall one problem that we have observed during the evaluation in \cite{nolle2016unsupervised}.
Whenever an activity is skipped or reworked, the remaining subsequence is shifted by one to the left, or the right respectively.
In \cite{nolle2016unsupervised} this led to the effect that all activities after the initial skipped (or reworked) activity had high reproduction error.
This phenomenon does not occur as severely in the extended approach, but it is still noticeable.
We assume that the additional hidden layers provide enough abstraction so the DAE can adapt to this problem.

Overall, we can see that the approach is very precise in narrowing down the exact cause of the anomaly.
In fact, this approach can be used to perform an automatic root cause analysis on the detected anomalies, without the need of an extra processing step.
Most other anomaly detection algorithms can only be used to divide the normal examples from the anomalies, but then an additional algorithm has to be used to, for instance, cluster the anomalies.
Another important point about this is that it allows to follow what the DAE has learned as well as to interpret it.
Usually, not being interpretable is a notorious problem for neural network based approaches.
Not in this case.

\subsection{Discussion}
At last we want to point out some interesting observations.
You might notice that in the two \textit{Skipping} examples the user at index 5, Amanda, produces a high reproduction error.
This is due to the fact that this event has a long-term dependency to an earlier event.
In the first case the event is connected to the \textit{PO Decreased} event, in the second one the connected event is the event that has been skipped.
Now that the trace has, in part, been shifted due to the skipping, the original event from index 6 is now at index 5.
Essentially, we have detected a fluke anomaly, that was not supposed to be there, yet the DAE approach has found it, demonstrating the feasibility of the approach.

This indicates, as already mentioned in \cite{nolle2016unsupervised}, that the DAE is sensitive towards the actual position of an event within a trace, which also becomes apparent in the second \textit{Skipping} example.
The event \textit{PO Created} is wrong, yet the DAE reproduces it correctly.
This is due to the fact the \textit{PO Created} can be correct here when the trace starts with the \textit{SC Created} event.

Furthermore, we still observe some cross-talk between adjacent events.
If we inspect the first \textit{Switching} example, we notice that \textit{SC Approved} and \textit{SC Purchased} have been switched, as correctly identified by the DAE.
However, the first event also produces a high reproduction error, albeit being correct at that location.
This error, compared to the error at indices 2 and 3, is significantly lower.

Table~\ref{tbl:f1_for_transition} provides the average $F_1$ score of the approaches when classifying traces, events, and attributes respectively.
When classifying attributes we classify the activity and the user separately, whereas when classifying events we do not separate the attributes.
Therefore, we also produced labels indicating anomalous and normal event attributes, when generating the datasets.
Any event attribute that had not been affected by any of the anomalies, has been labeled normal, whereas all other attributes have been labeled as anomalous.
An event is an anomalous when any of its attributes is anomalous and similarly a trace is anomalous when any of its events is anomalous.
Consequently, we can now calculate the performance of the DAE based on single events.

Sampling, t-STIDE, and t-STIDE+ can all also naturally be used to classify events.
Apart from DAE, only t-STIDE+, due to our adaption of the algorithm, can also be used to classify single attributes.
This can be done, for instance, by assigning the specific window anomaly scores to the respective last event or attribute in the window.
Sampling relies on a conformance check, which per definition gives a per event resolution.
Table~\ref{tbl:f1_for_transition} shows that the DAE outperforms the other approaches in all three categories.

\begin{table}
  \centering
  \caption{Results of the experiments for the anomalous event classifier per label and process model; best results are shown in bold typeface}
  \label{tbl:f1_for_transition}
  \begin{tabular}{llccc}
    \toprule
    Resolution & Method                                & Average         & Normal          & Anomaly         \\
    \midrule
    Traces     & Baseline                              & 0.44 $\pm$ 0.01 & 0.25 $\pm$ 0.01 & 0.62 $\pm$ 0.01 \\
               & HMM~\cite{warrender1999detecting}     & 0.46 $\pm$ 0.02 & 0.17 $\pm$ 0.06 & 0.75 $\pm$ 0.05 \\
               & OC-SVM~\cite{scholkopf1999support}    & 0.70 $\pm$ 0.06 & 0.51 $\pm$ 0.09 & 0.89 $\pm$ 0.05 \\
               & Naive~\cite{bezerra2013algorithms}    & 0.62 $\pm$ 0.02 & 0.49 $\pm$ 0.02 & 0.74 $\pm$ 0.02 \\
               & Sampling~\cite{bezerra2013algorithms} & 0.48 $\pm$ 0.05 & 0.11 $\pm$ 0.18 & 0.86 $\pm$ 0.09 \\
               & t-STIDE~\cite{warrender1999detecting} & 0.62 $\pm$ 0.02 & 0.49 $\pm$ 0.02 & 0.74 $\pm$ 0.02 \\
               & Likelihood~\cite{bohmer2016multi}     & 0.72 $\pm$ 0.15 & 0.52 $\pm$ 0.26 & 0.91 $\pm$ 0.07 \\
               & t-STIDE+                              & 0.83 $\pm$ 0.08 & 0.73 $\pm$ 0.12 & 0.93 $\pm$ 0.05 \\
               & DAE                                   & \textbf{0.87 $\pm$ 0.08} & \textbf{0.78 $\pm$ 0.13} & \textbf{0.95 $\pm$ 0.03} \\
    \midrule
    Events     & Sampling~\cite{bezerra2013algorithms} & 0.44 $\pm$ 0.17 & 0.64 $\pm$ 0.20 & 0.25 $\pm$ 0.14 \\
               & t-STIDE~\cite{warrender1999detecting} & 0.66 $\pm$ 0.03 & 0.90 $\pm$ 0.02 & 0.42 $\pm$ 0.04 \\
               & t-STIDE+                              & 0.61 $\pm$ 0.03 & 0.86 $\pm$ 0.03 & 0.36 $\pm$ 0.05 \\
               & DAE                                   & \textbf{0.72 $\pm$ 0.02} & \textbf{0.92 $\pm$ 0.02} & \textbf{0.53 $\pm$ 0.04} \\
    \midrule
    Attributes & t-STIDE+                              & 0.59 $\pm$ 0.03 & 0.87 $\pm$ 0.03 & 0.31 $\pm$ 0.06 \\
               & DAE                                   & \textbf{0.70 $\pm$ 0.03} & \textbf{0.94 $\pm$ 0.01} & \textbf{0.47 $\pm$ 0.05} \\
    \bottomrule
  \end{tabular}
\end{table}

We can conclude that this approach can discover the special characteristics of anomalies in an otherwise unknown process, while still being able to correctly identify normal behavior.
All together, we can say that the DAE approach is the most versatile out of all the approaches, as it works well in all of our test settings.

\section{Conclusion}
\label{sec:conclusion}

We have presented a novel application of denoising autoencoders to detect anomalies in business process data.
Our approach does not rely on any prior knowledge about the process itself.
Also, we do not rely on a clean dataset for the training; our approach is trained on a noisy dataset already containing the anomalies.
Furthermore, we have demonstrated that the autoencoder can also be used to easily identify the anomalous event(s) or event attribute(s), making results interpretable with regards to why an anomaly has been classified as such.
Even though we showed that this approach works for business process data, it can be applied just as easily to other domains with discrete sequential data.

We conducted a comprehensive evaluation using representative artificial and real-life event logs.
These event logs featured a range of different anomalies, different complexities in terms of the process model, variable variant probabilities, random user sets for each activity, and different shares of anomalous traces, ranging from 10\% to 100\%.
We compared the autoencoder approach to 7 other state-of-the-art anomaly detection methods, as described in \cite{chandola2012survey,bezerra2013algorithms,bohmer2016multi,warrender1999detecting}, showing that our approach outperforms all other methods in all of the test settings, reaching an $F_1$ score of 0.87 on average, whereas the second-best approach, our own adaption of the t-STIDE algorithm reached 0.83.
The next best unaltered anomaly detection algorithm, using an extended likelihood graph, reached an $F_1$ score of 0.72.
To our knowledge, this is the most sophisticated evaluation and comparison of anomaly detection methodology within the domain of process intelligence to date.

The biggest advantage of the autoencoder approach over the other methods is that it allows to analyze the detected anomalies even further.
Computing the anomaly score for each event attribute individually, the approach indicates the anomalous attribute very convincingly.
To our knowledge, this method of analyzing the anomalies is novel to the field of discovery science, as well as business intelligence and process mining.

The presented approach is an extended version of the approach from \cite{nolle2016unsupervised}.
In the original paper, we postulated that the approach is susceptible to anomalous behavior in the event log that is very frequent.
However, by showing that the approach works well for all noise levels, especially the higher noise levels where the exact same anomaly can occur many times, we have shown this not to be the case.
We also showed, by using skewed variant distributions, that the autoencoder is robust towards process models with unequally distributed variants, that is, some variants (i.e., one valid path through the process model) are more likely than others.
By including the user as an event attribute, we demonstrated that more dimensions can be added easily to the approach, without a significant loss of accuracy.

As an inspiration for future work on the matter we want to give a few remarks.
Note that for the DAE approach to work in a real-time setting the trace length of all future traces must be conform with the input size of the neural network.
If traces surpass the input size, they cannot be fed into the autoencoder.
There are some strategies to compensate for this problem.
For example, the autoencoder can be set up with spare padding input units.
Instead of padding all traces to match the maximum length encountered in the training set, we pad all traces to an arbitrary length greater than the maximum length.
If we do want to reuse the already trained autoencoder, we can use another strategy.
Every trace that is too long to feed into the autoencoder is divided into subsequences of exactly the size of the input.
For example, if an autoencoder has input size 10 and a trace has size 12, we would first feed the sequence starting from the first event until the tenth event, then the sequence from the second until the eleventh, and so on.
Then we can average the anomaly scores over all subsequences.
Another solution to the problem is the use of recurrent neural networks, which can be used to consume sequences of arbitrary length.

Another problem arises if one of the attributes is set to a value not encountered during training.
Consequently, there will be no dimension allocated in the one-hot encoding for it.
A simple solution to this problem is to add one extra dimension to the encoding vector which is used to encode all unknown attribute characteristics.
Nevertheless, the autoencoder should be retrained regularly to counteract concept drift.

With t-STIDE+ and DAE we have presented two approaches to detect anomalies in business process data.
It is quite costly to train a neural network on big datasets, because the dataset needs to be iterated many times.
The t-STIDE+ approach has the advantage that it can be trained with just one iteration.

However, due to the nature of the algorithm, it has some drawbacks; it cannot capture long-term dependencies if the window size is too small and if the windows size is too big the accuracy decreases.
Furthermore, it is not trivial to assign an anomaly score to a single attribute of an event, because anomaly scores are based on windows.
Lastly, it cannot deal with numerical event attributes (e.g., prices), without resorting to binning or grouping, which is not obvious.

The DAE approach does not have these drawbacks.
Numerical data can easily be modelled by using a single linear input and output neuron for real-valued numbers.
Certainly, it does require more training time, but with the introduction of evermore powerful GPUs and lately TPUs the trade-off between accuracy and efficiency is not as severe.

Overall, the results presented in this paper suggest that a denoising autoencoder is a reliable and versatile method for detecting---and interpreting---anomalies in unknown business processes.

\begin{acknowledgements}
This project (HA project no. 522/17-04) is funded in the framework of Hessen ModellProjekte, financed with funds of LOEWE – Landes-Offensive zur Entwicklung Wissenschaftlich-\"okonomischer Exzellenz, F\"orderlinie 3: KMU-Verbund-vorhaben (State Offensive for the Development of Scientific and Economic Excellence).
\end{acknowledgements}

\newpage

\bibliographystyle{spmpsci}         
\bibliography{references}           

\end{document}